\documentclass[letterpaper]{article} % DO NOT CHANGE THIS
\usepackage[preprint]{aaai2027}  % Open-source/preprint version.
\usepackage[hyphens]{url}  % DO NOT CHANGE THIS
\usepackage{graphicx} % DO NOT CHANGE THIS
\usepackage{natbib}  % DO NOT CHANGE THIS AND DO NOT ADD ANY OPTIONS TO IT
\usepackage{caption} % DO NOT CHANGE THIS AND DO NOT ADD ANY OPTIONS TO IT
\usepackage{algorithm}
\usepackage{algorithmic}
\usepackage{booktabs}
\usepackage{amsmath}
\usepackage{amssymb}
\usepackage{multirow}
\usepackage{array}
\usepackage{pgfplots}
\pgfplotsset{compat=1.18}
\usetikzlibrary{patterns}

\title{ReflectWorld-MM: An Entity-Oriented Multimodal Memory System for Open-Ended Video Streams}

\author{
    Xiaokang Ma\textsuperscript{1},
    Yifan Sun\textsuperscript{2,*},
    Zhihong Jin\textsuperscript{3,*},
    Jie Gu\textsuperscript{1,\dag},
    Yudong Luo\textsuperscript{1},
    Shenyi Shao\textsuperscript{1}\\
    Chu Tang\textsuperscript{1},
    Jingmin Chen\textsuperscript{1},
    Li Pu\textsuperscript{1}
}
\affiliations{
    \textsuperscript{1}Rightly Robotics\\
    \textsuperscript{2}Hangzhou Institute for Advanced Study, UCAS\\
    \textsuperscript{3}Zhejiang University
}

\begin{document}

\maketitle
\begingroup
\renewcommand{\thefootnote}{}
\footnotetext{\textsuperscript{*}Interns at Rightly Robotics. \quad
\textsuperscript{\dag}Corresponding author.}
\endgroup

\begin{abstract}
Building assistants that can continually watch the world, remember what they see,
and reason over their accumulated experience is a long-standing goal, and recently
multimodal agents equipped with long-term memory over video streams have attracted
increasing interest. Unfortunately, existing systems either keep their memory
inside the model context or in a flat feature store, and organize it around frames
rather than around the persistent entities a stream is really about, which confines
them to bounded videos and weakens their ability to track who and what reappears
over time. In this paper, we propose \emph{ReflectWorld-MM}, an entity-oriented
multimodal memory system for open-ended video streams. It consists of three parts.
The first is a perception front-end that turns an audiovisual stream into
entity-resolved observations under a bounded short-term memory. The second is a
hierarchical long-term memory, grounded in human memory theory, that couples a
multi-scale episodic memory, an evolving entity-centric semantic memory, and a
procedural memory. The third is a complete realization, built for real-world
operation, that ingests arbitrary streams and plugs into off-the-shelf assistants.
Across six long-video and lifelong-memory benchmarks, ReflectWorld-MM achieves the
best accuracy on all six, outperforming strong memory agents and a frontier model.
\end{abstract}

% =====================================================================
\section{Introduction}
% =====================================================================
\begin{figure}[t]
\centering
\includegraphics[width=0.85\columnwidth]{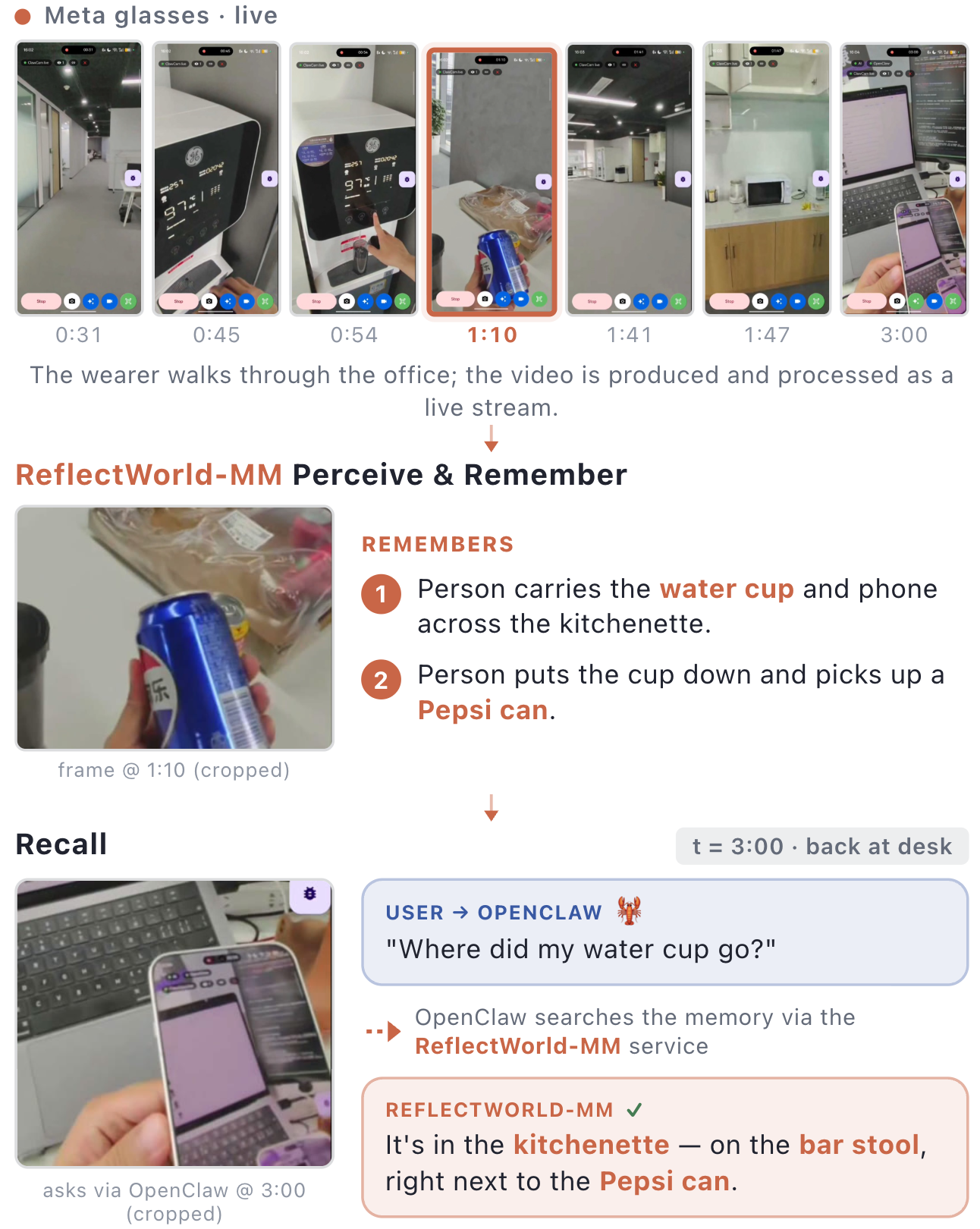}
\caption{From egocentric video to entity-oriented multimodal memory. A person wearing Meta
glasses wanders an office, producing a live video stream. ReflectWorld-MM perceives the stream
continuously and writes memory. The OpenClaw assistant can then answer the question about the
water cup by retrieving the matching memory.}
\label{fig:teaser}
\end{figure}

An intelligent assistant that keeps watching the world should also be able to
remember it. Cameras are now everywhere, from wearable glasses to household robots
and smartphones. There is thus a growing need for systems that perceive a continuous
multimodal stream, accumulate what they observe into a long-term memory, and answer
questions or act upon that memory at any later time. This is the difference between a
video-understanding model, which reasons over one clip, and an assistant that must
recall who appeared yesterday and how a situation has changed since.

A large proportion of current approaches store this memory implicitly, inside the
model. Streaming and long-video models compress frames into a memory bank, a sparse
token cache, or a key--value cache that is read back during
decoding~\citep{malmm,moviechat,flashvstream,rekv,videollmonline}. These methods are
effective on videos of bounded length, but their memory is single-scale and
content-agnostic, organized by frame or token rather than by entity and kept in the
model context or in main memory. Accordingly, it degrades once the stream grows far
beyond the videos the model was tuned on. In contrast, a separate line of work studies
explicit memory for language agents, and decomposes it into episodic, semantic, and
procedural components in the spirit of cognitive
theory~\citep{memgpt,generativeagents,amem,mem0,memorybank}. Unfortunately, these
systems are text-only and conversation-driven: they neither perceive a video stream nor
resolve the persistent visual entities needed to organize observations consistently across time.

The closest works to ours equip a multimodal agent with an explicit long-term memory
over video. M3-Agent organizes memory as an entity-centric multimodal graph and reasons
over it with a multi-turn retrieval policy~\citep{m3agent}, while WorldMM builds
episodic, semantic, and visual memories at multiple temporal scales for long-video
question answering~\citep{worldmm}. These are important steps, yet several gaps remain.
M3-Agent updates its store mainly by a frequency-based weight vote, so its semantic
knowledge is largely append-only and cannot be revised or removed once written. WorldMM
hand-sets its temporal scales per dataset and does not maintain entity identities, so it
cannot answer questions that hinge on who a person is. More fundamentally, neither
system unifies a multi-scale episodic memory, a continuously evolving entity-centric
semantic memory, and a procedural memory in one architecture, and neither is realized as
a complete service that runs on an arbitrary live stream.

In this paper, we propose \emph{ReflectWorld-MM}, an entity-oriented multimodal
memory system for open-ended video streams (Figure~\ref{fig:teaser}), built around
three components. First, a dedicated perception front-end converts a raw stream into
entity-resolved observations. Before each segment is interpreted, the model receives
working memory, scene context, entity history, and agent-provided scene understanding,
so the past participates in seeing the present rather than only being retrieved later.
Second, these observations are written into an entity-oriented memory that combines
multi-scale episodic memory, evolving semantic memory, and procedural memory for user
rules. This gives the system both event-level continuity and entity-level knowledge
that can support later question answering and proactive response. Third,
ReflectWorld-MM externalizes memory into a persistent, indexed service, keeping
per-segment cost bounded while supporting arbitrary video sources and off-the-shelf
assistants.

The main contributions of this paper are as follows:
\begin{itemize}
\item We propose a complete, hierarchical, entity-oriented multimodal memory
architecture grounded in human memory theory~\citep{tulving1985}. Both episodic and
semantic memory are organized around persistent entities: episodic memory links
observations through entity, trace, and schema levels, while semantic memory maintains
evolving entity-level knowledge over time.
Together with procedural memory for user rules and immediate response, these stores
provide an integrated memory substrate for open-ended video streams.
\item We design a perception front-end in which \emph{the past participates in seeing the
present}. Each segment is re-identified into entities and interpreted by a
vision--language model whose prompt is enhanced with within-event working memory,
per-camera scene context, and the retrieved history of any recognized entity. A
high-level agent also injects its scene understanding into the prompt to steer what the
model attends to before memory is written, giving perception narrative continuity along
an open-ended stream.
\item We evaluate ReflectWorld-MM on six long-video benchmarks, where it achieves strong
performance and outperforms prior memory competitors. Beyond accuracy, ReflectWorld-MM
shows a favorable answer-efficiency profile, maintaining rare video fallback while
achieving high accuracy. Finally, ReflectWorld-MM is implemented as a database-backed service with
video-source adapters, an interactive dashboard, and agent-facing APIs, allowing arbitrary
video streams to be ingested and arbitrary agentic systems to query the same memory
backend.
\end{itemize}

To the best of our knowledge, ReflectWorld-MM is the first system to combine
entity-oriented multimodal memory with open-ended streaming perception, cognitively
grounded hierarchical organization, and real-world deployment. The code of our system
has been open-sourced.

% =====================================================================
\section{Related Work}
% =====================================================================
\subsection{Streaming and Long-Video Memory}
A large body of work extends video models to long inputs by equipping them with a
memory. MovieChat merges dense tokens into a sparse memory for hour-long
videos~\citep{moviechat}, and MA-LMM compresses frames into a memory bank to bypass
the context limit~\citep{malmm}. Flash-VStream and VideoLLM-online target real-time
streams, but their memory is a compact feature or token state read by the
model~\citep{flashvstream,videollmonline}, and ReKV offloads and retrieves key--value
caches for streaming question answering~\citep{rekv}. A related group of agentic
methods selects query-relevant frames or builds a per-video tree at inference
time~\citep{videoagent_wang,videoagent_fan,videotree}. Nevertheless, their memory is
content-agnostic and organized by frame or token, and it lives inside the model or in
main memory. In contrast,
our system organizes memory around entities, externalizes it into a structured
store, and is thus not tied to a single bounded video.

\subsection{Memory for Language Agents}
Explicit memory has been studied for language agents. Mem-GPT pages information
between the context and an external tier~\citep{memgpt},
while Generative Agents maintain a memory stream with retrieval and
reflection~\citep{generativeagents}. More recent systems refine how memory is written
and updated, including the agentic notes of A-MEM, the extract-and-update pipeline of
Mem0, and the forgetting curve of MemoryBank~\citep{amem,mem0,memorybank}. These
systems decompose memory into episodic, semantic, and procedural components in the
spirit of cognitive theory. Unfortunately, they are text-only and conversation-driven:
they neither perceive video streams nor resolve persistent visual entities across time.
ReflectWorld-MM extends this explicit, cognitively grounded view of memory to streaming
multimodal perception.

\subsection{Entity-Centric and Lifelong Multimodal Memory}
The works closest to ours give a multimodal agent an explicit long-term memory.
M3-Agent stores an entity-centric multimodal graph and reasons over it with multi-turn
retrieval~\citep{m3agent}, and WorldMM builds multi-scale episodic, semantic, and
visual memories for long-video reasoning~\citep{worldmm}. On the text side,
knowledge-graph memories such as HippoRAG and temporal graphs such as Zep model
entities and their evolving relations~\citep{hipporag,zep}, GraphRAG summarizes a
corpus through an entity graph~\citep{graphrag}, and egocentric assistants and
benchmarks push toward lifelong, in-the-wild
perception~\citep{ego4d,egolife,teleego}. Our approach differs from these works in
mainly two points. First, ReflectWorld-MM unifies a multi-scale episodic memory, an
evolving and editable entity-centric semantic memory, and a procedural memory, while
keeping events and durable knowledge in separate indexed stores that can be edited
independently rather than coupled into a single multimodal graph. Second,
ReflectWorld-MM is realized as a system that operates over open-ended real-world
streams rather than over a fixed recording.

% =====================================================================
\section{ReflectWorld-MM}
% =====================================================================
\subsection{Problem Setup and Overview}
We consider an open-ended multimodal stream comprising a possibly unbounded sequence
of timestamped visual frames and audio from a live source. At any time $T$, the
system may receive a natural-language query $q$ and must answer it using everything
observed up to $T$. To this end, it maintains a long-term memory $\mathcal{M}$ that is
updated online as the stream arrives. We organize $\mathcal{M}$ around \emph{entities}.
An entity is a person or object that the system detects and re-identifies across time
under a persistent identifier. An entity carries both the events it participates in and
the durable knowledge accumulated about it. Because the stream has no predetermined
end, the system must run within a per-step cost that does not grow with how long it has
been running.

\begin{figure*}[t]
\centering
\includegraphics[width=0.95\textwidth]{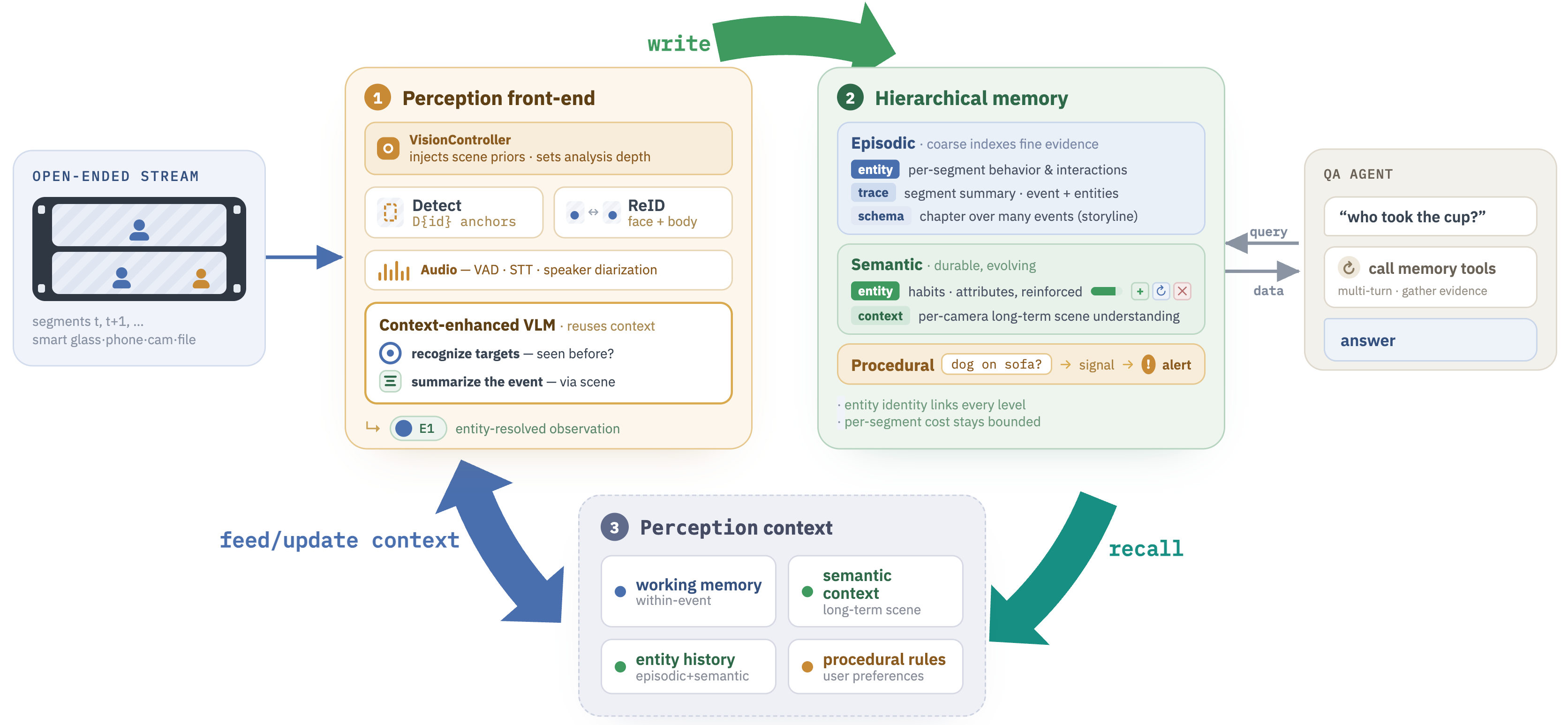}
\caption{System architecture of ReflectWorld-MM. The perception front-end converts an
open-ended stream into entity-resolved observations, writes them into hierarchical
episodic, semantic, and procedural memory, and feeds retrieved context back into
perception; a QA agent queries the same memory through tools.}
\label{fig:arch}
\end{figure*}

Figure~\ref{fig:arch} and Algorithm~\ref{alg:online} summarize the system. A perception
front-end ingests the stream, cuts it into activity-coherent segments, grounds a
vision--language model on local detection and re-identification evidence, and emits
entity-resolved observations. These observations are written into a hierarchical
long-term memory of three kinds---episodic, semantic, and procedural---following the
classical taxonomy of human memory~\citep{tulving1985}. The whole memory is externalized
into a persistent, indexed store, so the system can run indefinitely and be queried by an
assistant at any time. We detail each component below.

Three principles guide the design. First, we separate evidence from decisions. Detectors and
re-identifiers only produce scored evidence, and a single resolver owns every entity
identity decision, which keeps errors local and auditable. Second, we organize memory by
time scale and abstraction level rather than as a flat log, letting the agent retrieve at
the granularity a question needs. Third,
the past participates in perceiving the present: before the model interprets a segment, we
enhance it with the accumulated context, so that ReflectWorld-MM actively makes sense of
each scene and understands the world continuously, rather than describing segments in
isolation and using the past only when later answering a query.

\begin{algorithm}[t]
\caption{Online memory construction in ReflectWorld-MM}
\label{alg:online}
\begin{algorithmic}[1]
\REQUIRE stream $\mathcal{S}$; consolidation interval $N$; schema budget $B$
\STATE init short-term memory $\mathcal{W}\!\gets\!\varnothing$, store $\mathcal{M}\!\gets\!\varnothing$
\FOR{each segment $s$ in $\textsc{Segment}(\mathcal{S})$}
  \STATE $D\gets\textsc{Detect}(s)$;\ \ $\mathit{Ev}\gets\textsc{ReID}(D)$ \COMMENT{local evidence}
  \STATE $c\gets\textsc{Context}(\mathcal{W},\mathcal{M},\mathit{Ev})$ \COMMENT{working memory + scene + entity history}
  \STATE $(\ell,\text{prompt})\gets\textsc{Steer}(s,c)$ \COMMENT{agent-steered, context-enhanced}
  \STATE $O\gets\textsc{VLM}(s,\text{prompt},\ell)$ anchored to $D,\mathit{Ev}$
  \STATE $E\gets\textsc{Resolve}(O,\mathit{Ev})$ \COMMENT{evidence $\to$ identity}
  \STATE $\mathcal{W}\gets\textsc{UpdateShortTerm}(\mathcal{W},O,E)$ \COMMENT{narrative continuity}
  \STATE $\mathcal{M}.\textsc{WriteEntity}(E)$;\ \ $\mathcal{M}.\textsc{WriteTrace}(O,E)$
  \FORALL{entity $e$ updated in $E$}
    \IF{$\textsc{Count}(e)\bmod N=0$}
      \STATE $\mathcal{M}.\textsc{Consolidate}(e)$ \COMMENT{\textsc{Add}/\textsc{Update}/\textsc{Delete} + reinforce}
    \ENDIF
  \ENDFOR
  \IF{event closed \textbf{and} ($\#\text{events}\ge B$ \textbf{or} interval elapsed)}
    \STATE $\mathcal{M}.\textsc{WriteSchema}(\text{chapter})$
  \ENDIF
\ENDFOR
\end{algorithmic}
\end{algorithm}

\subsection{Perception Front-End}
The perception front-end converts a raw stream into entity-resolved observations. We
describe it by its four parts.

\noindent\textbf{Streaming ingestion.} The system accepts arbitrary live sources through a
common gateway, described in Appendix~\ref{app:streams}. The gateway normalizes each
source and splits it into activity-coherent audiovisual segments with transcripts,
including speaker diarization when available. As a result, the front-end operates on an
open-ended stream instead of a pre-cut clip, and it does not assume the video has an end.

\noindent\textbf{Entity recognition.} Within each segment, the front-end detects the
persons and objects present and re-identifies them against the entities it has already
seen. For persons, identity evidence comes from both face and body appearance, which is
matched against per-entity galleries so that the same person keeps a stable identifier
whenever they reappear, even across different sessions. Following our first design
principle, re-identification only produces evidence: it proposes candidate matches with
similarity scores. A dedicated resolver then makes the final identity decision and is
the only component allowed to write an identity. This separation keeps a single noisy
match from corrupting the memory, and it makes every identity decision auditable.

\noindent\textbf{Context-enhanced perception.} A defining feature of ReflectWorld-MM is that
the model does not interpret a segment in isolation: before it runs, the segment's prompt is
enhanced with the relevant context the system has already accumulated, so that the past
participates in seeing the present. Three layers of context are assembled. The first is a
bounded \emph{working memory} that maintains the current event state and gives perception
within-event narrative continuity, so that a person who briefly leaves the frame is still
read as the same ongoing actor. Its concrete bounds are given in
Appendix~\ref{app:implementation}. The second layer is a per-camera \emph{semantic context}
that supplies cross-event scene
background---what kind of place this is and its typical activities and routines---so the
model interprets the segment against the scene rather than from scratch. The third is
\emph{entity history}: once re-identification recognizes an entity, the system retrieves that
entity's accumulated episodic and semantic knowledge and adds it to the prompt, so the
current frame is understood in light of who and what is present. This perception-time use of
context is a key difference from prior multimodal agents, which store memory first and
consult the past only when later answering a query.

\noindent\textbf{Agent-steered adaptive perception.} Perception is not a fixed extractor; it
is actively steered by a high-level agent that makes sense of the scene. The agent maintains
an understanding of what the camera is looking at and what currently matters, and it injects
this understanding---a scene description, the targets to focus on, and any active user
rules---into the perception prompt, which redirects where the model attends and what it is
asked to report. The agent also decides how much computation each segment receives: a
controller with a deterministic frame-level policy and a semantic segment-level policy
decides whether a segment is analyzed at all and how richly, so that static or silent
segments are handled cheaply while information-rich or rule-relevant ones receive a fuller
analysis. Hard guards and detection anchoring prevent the scheduler from skipping critical
evidence or writing ungrounded identity targets; the exact policies are in
Appendix~\ref{app:implementation}.

\subsection{Hierarchical Episodic Memory}
The episodic memory records what happened, and it is organized at three levels of
increasing abstraction, from entity observations to trace summaries and schema-level
chapters. This hierarchy follows the autobiographical structure of human
memory~\citep{conway2000,conway2005} and lets coarse memories index fine-grained events.

\noindent\textbf{Entity level.} The finest level stores one observation per entity per
segment. It records a single person or object together with its appearance, its
behavior, and its interactions at that moment, and it is keyed by the entity's
persistent identifier. This level corresponds to event-specific knowledge, the most
concrete layer, and it is what lets the system later recall what a
particular entity was doing at a particular time.

\noindent\textbf{Trace level.} The middle level stores one summary per segment. It
describes the event that took place and lists the entities involved, linking back to
their entity-level observations. The trace level is the default unit of recall, and it
ties together the otherwise separate per-entity observations of the same moment.

\noindent\textbf{Schema level.} The coarsest level stores a narrative chapter that
aggregates many segments into a storyline. When an event closes, the system accumulates
it, and once a budget of events or a time interval is reached, it consolidates the
accumulated events into a chapter-level summary. This compact index bounds how many
fine-grained traces a query must consider when navigating a long history; additional
write details are given in Appendix~\ref{app:implementation}.

\subsection{Evolving Entity-Centric Semantic Memory}
Beyond individual events, ReflectWorld-MM distills durable knowledge about each entity,
such as attributes, habits, and relations. This semantic memory is consolidated after
every $N$ new observations of an entity. The consolidator reads that entity's prior
episodic and semantic records and emits one of four edit decisions: \textsc{Add},
\textsc{Update}, \textsc{Delete}, or no change. This makes the memory evolving rather
than append-only: new evidence can reinforce an existing fact, revise a stale one, or
delete a fact that is no longer supported. Safe edit targeting and identity-fact guards
are implementation details; Appendix~\ref{app:implementation} provides the exact
consolidation schedule and safeguards.

Each semantic fact carries an importance score that grows when the fact is confirmed
again. Denoting the importance by $w\in[0,1]$, a re-confirmation updates it as:
\begin{equation}
w \leftarrow w + (1-w)\,\gamma ,
\label{eq:reinforce}
\end{equation}
where $\gamma\in(0,1)$ is a growth rate. Equation~\eqref{eq:reinforce} increases $w$
asymptotically toward one, so repeated evidence makes a fact more stable while newly
added facts remain easier to revise. This is the main difference from prior multimodal
agents such as M3-Agent, whose memory is effectively append-only~\citep{m3agent}. We
therefore use a qualitative case to show how entity semantic memory changes and is
strengthened over time.

\subsection{Procedural Memory and Proactive Response}
The procedural memory stores rules and preferences supplied by the user, such as which
situations should raise an alert. These rules are surfaced into the perception prompt, so
the stream is analyzed only once. The vision--language model does not act on them directly;
instead it emits \emph{semantic signals}, for example that a situation looks risky or which
rules a segment appears to satisfy. Following the same evidence-versus-decision principle, a
separate deterministic resolver then combines these signals with the per-camera
notification policy and a short-term record of what was recently sent, and decides what, if
anything, to deliver. A rule match is therefore evidence that informs the decision, not a
trigger that forces it, which avoids repeated or spurious alerts.

\subsection{Retrieval and System Realization}
\noindent\textbf{Retrieval.} Queries are answered from the externalized store through
ordinary tool calls. The agent chooses the memory level to search---entity observations,
traces, schemas, semantic facts, or procedural rules---and retrieves only the items needed
for the question. Shared entity identifiers connect levels, so an event can recover its
participants and an entity can recover its history. The importance score participates in
ranking, favoring well-confirmed facts. Thus the agent attends over a retrieved subset
rather than the full stream history.

\noindent\textbf{System realization.} The memory is persisted in an indexed database rather
than held in model context. The total store naturally grows as the system remembers more,
but the per-segment working state and query context remain bounded. This is the practical
difference from streaming-cache methods such as ReKV~\citep{rekv}, which must still read a
state tied to stream length. The same interface makes ReflectWorld-MM deployable: it
ingests live sources through a common gateway and exposes capture, perception, memory
query, notification policy, and context management as tools. The benchmark agent and the
OpenClaw deployment share this interface. Implementation details for the stream gateway,
agent API, and low-latency commit path are given in Appendix~\ref{app:streams},
Appendix~\ref{app:agent}, and Appendix~\ref{app:implementation}, respectively.

% =====================================================================
\section{Experiments}
% =====================================================================
\subsection{Setup}
\noindent\textbf{Benchmarks.} We evaluate on six benchmarks spanning general, egocentric,
and entity-centric settings: VideoMME-Long, LVBench, HippoVlog, EgoLife-QA,
M3-bench-robot, and M3-bench-web. This mix checks whether ReflectWorld-MM preserves
ordinary long-video understanding while improving settings where identity and persistent
memory are central; detailed benchmark descriptions are given in
Appendix~\ref{app:benchmarks}.

\noindent\textbf{Baselines.} We compare against the two strongest multimodal memory
agents, M3-Agent~\citep{m3agent} and WorldMM~\citep{worldmm}, and against a frontier
model, GPT-5, as a reference for raw video understanding without a dedicated memory.

\noindent\textbf{Implementation.} Following WorldMM's evaluation setting for a fair
comparison, ReflectWorld-MM uses GPT-5-mini for memory extraction and semantic
consolidation, and GPT-5 as the agent that answers a query by retrieving from memory.
Persons are re-identified from face and body appearance, and every memory item is indexed
with OpenAI's \texttt{text-embedding-3-small}. Accuracy on the multiple-choice benchmarks is measured by option
matching, while the open-ended M3-bench answers are scored by a GPT-5-mini judge, since the
GPT-4o judge used by the original benchmark is no longer available. For fairness, we re-run
M3-Agent with its official source code on EgoLife-QA and M3-bench, re-run WorldMM with its
official code on VideoMME-Long, and run GPT-5 ourselves on EgoLife-QA and M3-bench. Other
baseline numbers are quoted from the corresponding papers. Appendix~\ref{app:provenance}
documents the provenance of the reported numbers.

\begin{table}[t]
\centering
\footnotesize
\setlength{\tabcolsep}{4.5pt}
\caption{Results on the entity-sensitive benchmarks: multiple-choice accuracy on EgoLife-QA
and GPT-judged correctness on the open-ended M3-bench. Best in \textbf{bold}.
Appendix~\ref{app:provenance} documents number provenance and re-run settings.}
\label{tab:main1}
\begin{tabular}{lccc}
\toprule
Method & EgoLife-QA & M3-robot & M3-web \\
\midrule
\multicolumn{4}{l}{\textit{Vision--language models}} \\
Qwen2.5-VL-7B          & --   & 3.4  & 14.9 \\
Gemini-1.5-Pro         & 36.9 & 8.0  & 23.2 \\
GPT-5                  & 42.6 & 34.7 & 53.9 \\
\midrule
\multicolumn{4}{l}{\textit{Long-video / online methods}} \\
MovieChat              & --   & 11.2 & 12.6 \\
Flash-VStream          & --   & 19.4 & 23.6 \\
\midrule
\multicolumn{4}{l}{\textit{Agent / memory methods}} \\
% Gemini-Agent           & --   & 16.9 & 34.1 \\
M3-Agent               & 30.8 & 28.3 & 45.6 \\
\midrule
ReflectWorld-MM        & \textbf{46.8} & \textbf{37.4} & \textbf{56.0} \\
\bottomrule
\end{tabular}
\end{table}

\subsection{Main Results}
We compare ReflectWorld-MM against a broad set of baselines. Table~\ref{tab:main1} covers
the three entity- and identity-sensitive benchmarks, and Table~\ref{tab:main2} covers the
three general long-video benchmarks. We have several observations. First, ReflectWorld-MM
achieves the best accuracy on all six benchmarks. Second, the largest gains appear on the
entity-centric M3-bench, where ReflectWorld-MM improves over the strongest prior method,
M3-Agent, by $10.4$ points on the web split and $9.1$ points on the robot split. We note,
however, that the absolute accuracies on M3-bench remain low for all systems, which
indicates that the benchmark is far from solved. Third, because ReflectWorld-MM uses GPT-5
as the agent that answers queries, the comparison against the GPT-5 reference isolates the
effect of the memory. Equipping the same model with ReflectWorld-MM improves long-video
question answering, and the improvement grows with video length, from $2.6$ points on
the shorter VideoMME to $9.0$ points on LVBench. Fourth, EgoLife-QA highlights the value
of entity identity. Its questions hinge on who a person is, and ReflectWorld-MM reaches
the best accuracy. Notably, the original EgoLife paper reports only $45.5$ accuracy even
when using manually annotated descriptions with identity IDs, whereas ReflectWorld-MM
reaches $46.8$ with automatically constructed memory. We do not report WorldMM in
Table~\ref{tab:main1} because it does not maintain explicit entity identities, which is
precisely the information these questions require. This is direct evidence for our
entity-oriented design.

\begin{table}[t]
\centering
\footnotesize
\setlength{\tabcolsep}{4pt}
\caption{Results on the general long-video benchmarks (accuracy, \%). Best in
\textbf{bold}.}
\label{tab:main2}
\begin{tabular}{lccc}
\toprule
Method & VideoMME-L & LVBench & HippoVlog \\
\midrule
\multicolumn{4}{l}{\textit{Base models}} \\
Qwen3-VL-8B            & 61.0 & 48.3 & 74.4 \\
Gemini 2.5 Pro         & 55.7 & 57.0 & 72.0 \\
GPT-5                  & 74.3 & 60.4 & 75.7 \\
\midrule
\multicolumn{4}{l}{\textit{Long-video LLMs}} \\
VideoChat-Flash        & 44.1 & 33.2 & 58.0 \\
Video-RTS              & 47.9 & 39.8 & 59.0 \\
\midrule
\multicolumn{4}{l}{\textit{RAG-based methods}} \\
LightRAG               & 46.6 & 30.4 & 47.4 \\
% HippoRAG               & 52.1 & 54.0 & 63.2 \\
Video-RAG              & 55.4 & 33.1 & 65.1 \\
\midrule
\multicolumn{4}{l}{\textit{Memory-based methods}} \\
HippoMM                & 41.6 & 38.2 & 71.9 \\
M3-Agent               & 55.3 & 49.3 & 65.5 \\
WorldMM                & 73.8 & 61.9 & 78.3 \\
\midrule
ReflectWorld-MM        & \textbf{76.9} & \textbf{69.4} & \textbf{80.9} \\
\bottomrule
\end{tabular}
\end{table}

\subsection{Memory Quality}
Accuracy alone does not reveal whether the stored memory is useful to an agent. A system
may answer with little context because its memory is sparse, or it may achieve high
accuracy only by repeatedly falling back to the source video. We therefore evaluate memory
quality through answer efficiency. Table~\ref{tab:eff} reports accuracy together with the
average number of answer tokens per question and the fraction of questions that still
require video fallback. The intended pattern is high accuracy with rare fallback: if the
memory contains the needed evidence, the agent should answer mostly from memory.

The results support this pattern. ReflectWorld-MM achieves the best accuracy with only
$4.6\%$ fallback on EgoLife-QA and $6.8\%$ on VideoMME-L, so nearly all questions are
answered from extracted memory rather than by reopening the video. The baselines clarify
why both accuracy and fallback matter: M3-Agent uses fewer tokens and $0.0\%$ fallback but
is much less accurate, indicating that compact traces miss needed evidence, while WorldMM
is strong on VideoMME-L but requires video for $34.0\%$ of questions. ReflectWorld-MM
therefore shows that its memory stores answerable evidence, not merely an index to the
source stream.

\begin{table}[t]
\centering
\footnotesize
\setlength{\tabcolsep}{4pt}
\caption{Memory quality measured by answer efficiency. For each benchmark we report
accuracy, average answer tokens per question (Tok.), and the percentage of questions that
still require video fallback (Vid.). Higher is better for accuracy; lower is better for
Tok. and Vid.}
\label{tab:eff}
\begin{tabular}{llccc}
\toprule
Benchmark & Method & Acc. & Tok. & Vid. \\
\midrule
EgoLife-QA & M3-Agent        & 30.8 & 13k & 0.0\% \\
           & ReflectWorld-MM & \textbf{46.8} & 55k & 4.6\% \\
\midrule
VideoMME-L & M3-Agent        & 55.3 & 8.5k  & 0.0\% \\
           & WorldMM         & 73.8   & 56k    & 34.0\% \\
           & ReflectWorld-MM & \textbf{76.9} & 43k & 6.8\% \\
\bottomrule
\end{tabular}
\end{table}

\begin{figure}[t]
\centering
\begin{tikzpicture}[x=1cm,y=1cm]
\def\barw{0.29}
\def\labelfont{\fontsize{5}{5}\selectfont}
\newcommand{\ablbar}[4]{%
  \path[#4] ({#1-\barw/2},0) rectangle ({#1+\barw/2},#2);
  \node[font=\labelfont, anchor=south] at (#1,{#2+0.06}) {#3};
}
\draw[black!65] (-0.20,0) -- (3.98,0);

\ablbar{0.00}{1.36}{37.4}{draw=blue!55!black, fill=blue!18, pattern=north east lines, pattern color=blue!55!black}
\ablbar{0.47}{0.98}{35.4}{draw=green!45!black, fill=green!25}
\ablbar{0.94}{0.66}{33.6}{draw=orange!70!black, fill=orange!35}
\ablbar{1.41}{0.82}{34.6}{draw=red!55!black, fill=red!25}
\ablbar{2.40}{1.46}{46.8}{draw=blue!55!black, fill=blue!18, pattern=north east lines, pattern color=blue!55!black}
\ablbar{2.87}{1.08}{45.9}{draw=green!45!black, fill=green!25}
\ablbar{3.34}{0.86}{45.6}{draw=orange!70!black, fill=orange!35}
\ablbar{3.81}{0.86}{45.6}{draw=red!55!black, fill=red!25}

\node[font=\scriptsize, anchor=north] at (0.705,-0.13) {M3-robot};
\node[font=\scriptsize, anchor=north] at (3.105,-0.13) {EgoLife-QA};

\draw[gray!45, rounded corners=2pt] (4.32,-0.08) rectangle (6.39,1.58);
\path[draw=blue!55!black, fill=blue!18, pattern=north east lines, pattern color=blue!55!black]
  (4.46,1.34) rectangle (4.61,1.49);
\node[font=\scriptsize, anchor=west] at (4.70,1.415) {Full};
\path[draw=green!45!black, fill=green!25] (4.46,0.96) rectangle (4.61,1.11);
\node[font=\scriptsize, anchor=west] at (4.70,1.035) {w/o entity};
\path[draw=orange!70!black, fill=orange!35] (4.46,0.58) rectangle (4.61,0.73);
\node[font=\scriptsize, anchor=west] at (4.70,0.655) {w/o schema};
\path[draw=red!55!black, fill=red!25] (4.46,0.20) rectangle (4.61,0.35);
\node[font=\scriptsize, anchor=west] at (4.70,0.275) {w/o semantic};
\end{tikzpicture}
\caption{Answer-time ablation on entity-sensitive benchmarks. In each group, left-to-right
bars denote Full, w/o entity, w/o schema, and w/o semantic; the Full bar is hatched.}
\label{fig:ablation}
\end{figure}
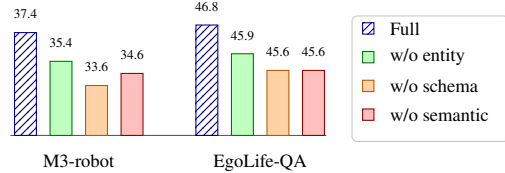

\begin{figure*}[!t]
\centering
\includegraphics[width=0.84\textwidth]{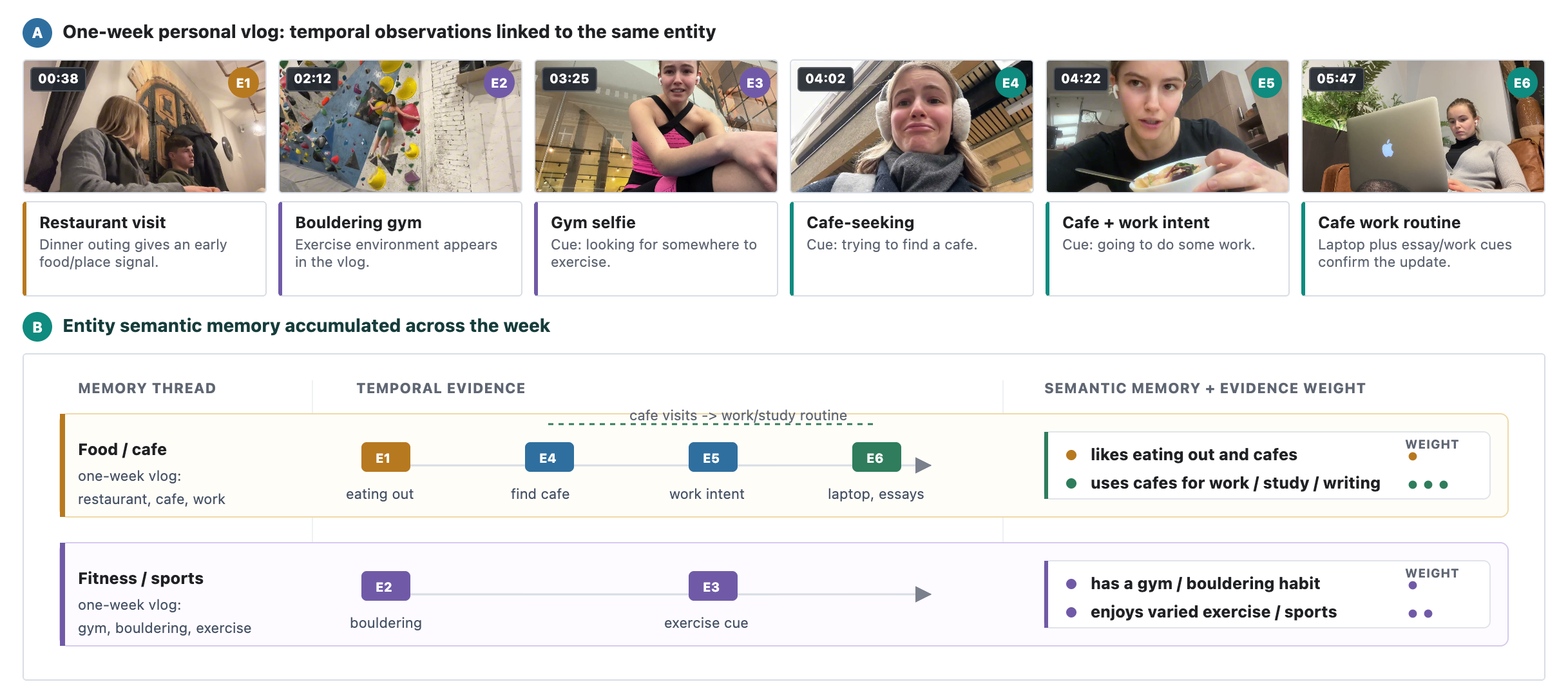}
\caption{Entity-centric semantic memory over a one-week vlog. Observations from different days are linked to the
same person, so repeated food/cafe and exercise evidence strengthens persistent semantic
facts.}
\label{fig:qual-entity}
\end{figure*}

\begin{figure*}[t]
\centering
\includegraphics[width=0.84\textwidth]{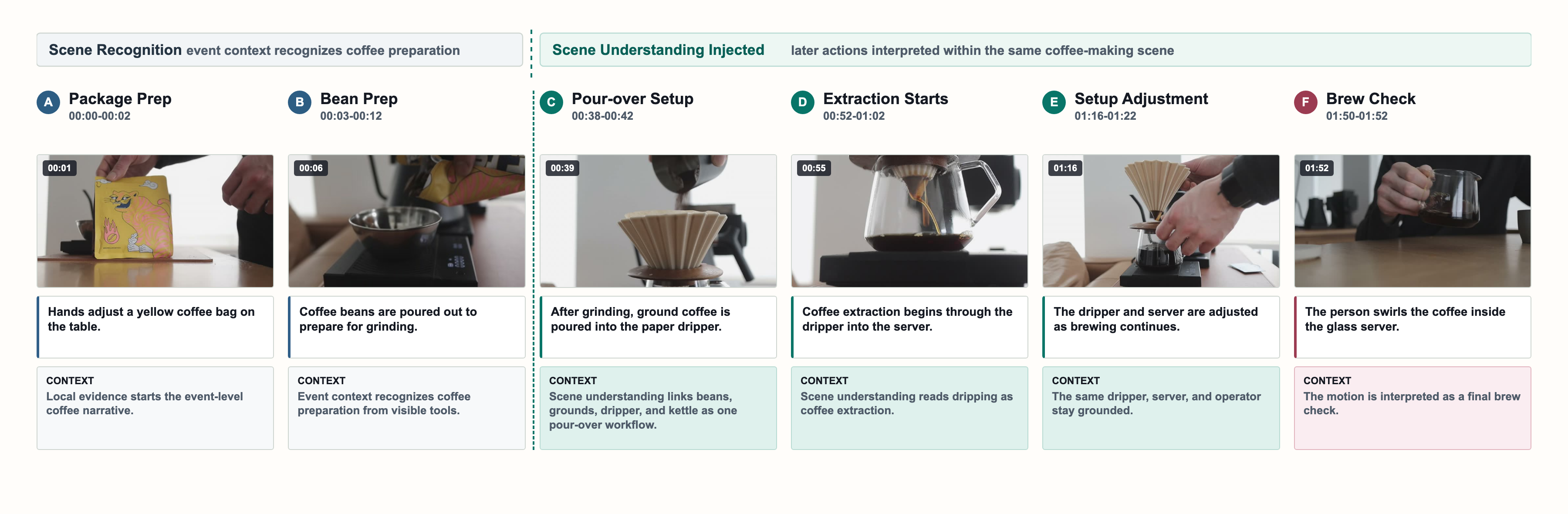}
\caption{Task-adaptive extraction in a coffee-making scene. Early coffee-making evidence sets the event context,
which guides later perception of dripper setup, extraction, adjustment, and checking as one
workflow.}
\label{fig:qual-adaptive}
\end{figure*}

\begin{figure}[!t]
\centering
\includegraphics[width=0.95\columnwidth]{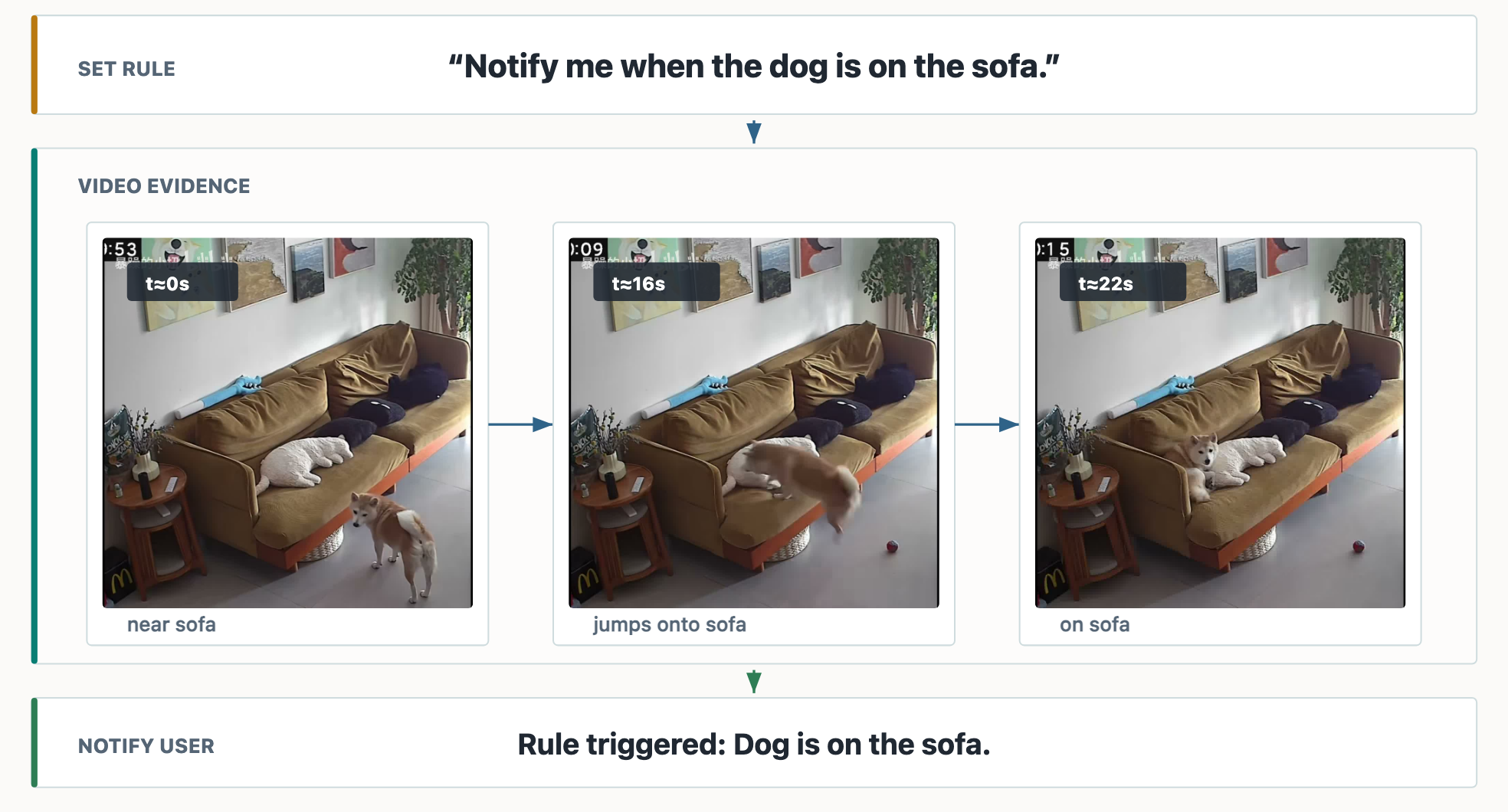}
\caption{Procedural memory for proactive notification. A stored dog-on-sofa rule,
grounded visual evidence, and deterministic resolution together trigger the notification.}
\label{fig:qual-procedural}
\end{figure}

\subsection{Ablation Study}
We ablate the most entity-sensitive benchmarks, M3-bench-robot and EgoLife-QA, because
they best match open-ended video streams where answers depend on following entities over
time. Rebuilding memory for every variant is costly, so we use an \emph{answer-time}
ablation: the full system builds memory once, and each variant blocks the answering agent
from using one component. Specifically, w/o entity association blocks entity-linked
retrieval, w/o schema removes the coarse event index, and w/o semantic hides distilled
entity knowledge.

Figure~\ref{fig:ablation} shows that this fixed-memory test is still diagnostic. Because
the stored memories are unchanged, each drop should be read as a lower bound on the
component's value: overlapping evidence can still be retrieved through other paths. Even
under this conservative setting, blocking a component and its related retrieval reduces
accuracy. The largest drop appears when schema memory is removed on M3-bench-robot
($37.4\rightarrow33.6$), while entity association and semantic memory also help on both
datasets. Together, these drops show that the measured gains depend on the proposed entity
association, schema-level episodic organization, and semantic consolidation, not only on the
answer model used at question time.

\subsection{Qualitative Analysis}
Aggregate scores do not show how memory is formed during perception. We therefore give
three case studies that expose the write side of ReflectWorld-MM.

\noindent\textbf{Entity-centric semantic memory.} Figure~\ref{fig:qual-entity} shows a
one-week personal vlog in which observations from different days are linked to the same
person. Because each moment is attached to the same entity, the semantic store can
consolidate them into durable facts rather than leave them as isolated clip summaries.
Specifically, food and cafe observations reinforce a work/study routine,
while the bouldering and gym observations reinforce an exercise habit. The example
illustrates why entity identity is not just a retrieval key: it is the unit over which
longitudinal evidence accumulates.

\noindent\textbf{Task-adaptive extraction.} Figure~\ref{fig:qual-adaptive} illustrates
context-enhanced perception on a coffee-making sequence. Once the system recognizes the
scene as coffee preparation, the scene understanding is
injected into later perception. Subsequent segments are then interpreted as stages of the
same pour-over workflow, including dripper setup, extraction, setup adjustment, and brew
checking. This case shows that the system does not extract a fixed list of generic fields
from every segment. Instead, accumulated scene context changes what later observations mean.

\noindent\textbf{Procedural memory and proactive response.} Figure~\ref{fig:qual-procedural}
shows how procedural memory turns perception into action. The user first sets the rule
``Notify me when the dog is on the sofa.'' During live perception, the system observes the
dog near the sofa, then jumping onto it, and finally lying on the sofa. The visual evidence
is not itself the action trigger; it is converted into a semantic signal and resolved
against the stored rule with policy, deduplication, and cooldown.

\subsection{Real-World Operation}
Beyond the benchmarks, ReflectWorld-MM runs as a complete service. The capture layer accepts
arbitrary video streams through a common gateway, including network cameras, webcams, local
files, HTTP streams, and smartphone cameras, and it continues processing open-ended video
without assuming a fixed duration. Its memory is persisted in an indexed database rather
than held inside the model, so any downstream agentic system can attach to the same
long-term memory. The code has been open-sourced; the stream gateway, agent interface, and
dashboard are described in Appendix~\ref{app:streams}, Appendix~\ref{app:agent}, and
Appendix~\ref{app:dashboard}. The benchmark agent, the dashboard in
Figure~\ref{fig:dashboard}, and OpenClaw-based agents all access the same deployed memory
backend rather than separate offline pipelines.

% =====================================================================
\section{Conclusion}
% =====================================================================
We presented ReflectWorld-MM, an entity-oriented multimodal memory system for
open-ended video streams. The system turns streaming audiovisual input into entity-resolved
observations and writes them into an externalized memory organized into episodic,
semantic, and procedural stores. This design makes memory persistent, revisable, and
accessible to downstream agents, while keeping perception grounded in both current visual
evidence and accumulated context. Across six long-video and lifelong-memory benchmarks,
ReflectWorld-MM achieves the strongest accuracy among the compared methods. The gains are
most pronounced in entity-centric settings, including M3-bench and EgoLife-QA, and the
answer-efficiency results show that the system usually answers from constructed memory
rather than by repeatedly reopening the source video. The ablation and qualitative cases
further indicate that entity association, schema-level abstraction, semantic
consolidation, and procedural rules contribute to different parts of the open-ended stream
problem. These results suggest that persistent multimodal memory should be treated as an
active system component: it participates in perception, organizes experience around
entities, and provides an interface through which general agents can recall and act over
long-running visual streams.

\clearpage
\raggedbottom

\bibliography{refs}

\clearpage

\appendix

\begin{center}
{\Large\bfseries Appendix}
\end{center}

\section{Provenance of Reported Numbers}
\label{app:provenance}
Table~\ref{tab:provenance} summarizes the provenance of the reported numbers. For
head-to-head comparisons on entity-sensitive benchmarks, we re-run M3-Agent with its
official source code and evaluate the resulting answers under the same protocol as
ReflectWorld-MM. For WorldMM, only the VideoMME-Long number is our re-run with the
official code; its LVBench and HippoVlog numbers are quoted from the WorldMM
paper~\citep{worldmm}. The GPT-5 reference row is our run on EgoLife-QA and M3-bench, and
is quoted from WorldMM for VideoMME-Long, LVBench, and HippoVlog.

M3-bench is open-ended and scored by an LLM judge. The original benchmark used a GPT-4o
judge, which is no longer available, so we evaluate GPT-5, M3-Agent, and our system with a
GPT-5-mini judge. The remaining M3-bench baselines are quoted under the original GPT-4o
judge and are marked accordingly.

\begin{center}
\footnotesize
\setlength{\tabcolsep}{3pt}
\captionof{table}{Provenance of reported evaluation numbers. Unless specified here, baseline
numbers are quoted from the corresponding original papers. MC denotes multiple-choice
accuracy computed by option matching.}
\label{tab:provenance}
\begin{tabular}{>{\raggedright\arraybackslash}p{0.31\columnwidth}
                >{\raggedright\arraybackslash}p{0.57\columnwidth}}
\toprule
Method & Provenance \\
\midrule
ReflectWorld-MM
& Our runs on all six benchmarks. Multiple-choice benchmarks use MC option matching;
M3-bench uses GPT-5-mini judging. \\
\midrule
GPT-5
& Our runs on EgoLife-QA and M3-bench. VideoMME-L, LVBench, and HippoVlog numbers are
quoted from WorldMM~\citep{worldmm}. \\
\midrule
M3-Agent
& Our re-runs with the official M3 source code on EgoLife-QA and M3-bench. The M3-bench
main-table numbers use GPT-5-mini judging; the detailed breakdown in
Table~\ref{tab:m3breakdown} quotes the M3 paper~\citep{m3agent}. \\
\midrule
WorldMM
& Our re-run with official WorldMM code on VideoMME-L. LVBench and HippoVlog numbers are
quoted from WorldMM~\citep{worldmm}. We do not report WorldMM on EgoLife-QA or M3-bench. \\
\midrule
Other baselines
& Quoted from the corresponding original papers, including M3-bench baselines judged with
the original GPT-4o judge and EgoLife-QA/long-video baselines under their reported
protocols. \\
\bottomrule
\end{tabular}
\end{center}

\section{Detailed Benchmark Breakdowns}
\label{app:benchmarks}
We evaluate on six benchmarks that cover complementary video-memory settings.
\emph{VideoMME} (long split)~\citep{videomme} and \emph{LVBench}~\citep{lvbench} are
general long-video, multiple-choice benchmarks, with videos up to about one and two hours
respectively; they test whether a memory system preserves ordinary long-video
understanding. \emph{HippoVlog}~\citep{hippomm} contains roughly one thousand
multiple-choice questions over audiovisual vlogs and probes memory formation and
associative recall across modalities. \emph{EgoLife-QA}~\citep{egolife} poses
life-oriented, often identity-dependent multiple-choice questions over a week-long
egocentric recording, through sub-tasks such as entity logging, event recall, and habit
insight. \emph{M3-bench}~\citep{m3agent} evaluates open-ended question answering over an
entity-centric memory, with a robot split of egocentric videos and a web split of online
videos, each averaging about half an hour; its questions require multi-hop, cross-modal,
and person-centric reasoning.

Table~\ref{tab:m3breakdown} gives the per-category breakdown on M3-bench, and
Table~\ref{tab:egobreakdown} the per-sub-task breakdown on EgoLife-QA. Baseline numbers
from earlier systems are quoted from the original papers; GPT-5 and ReflectWorld-MM rows
are produced by our current evaluation protocol.

\begin{table*}[t]
\centering
\footnotesize
\setlength{\tabcolsep}{4pt}
\caption{M3-bench per-category breakdown (\%). Rows through M3-Agent are quoted from
the M3 paper~\citep{m3agent} under its original GPT-4o judge. GPT-5 and ReflectWorld-MM
are our re-runs with a GPT-5-mini judge. ME: multi-evidence, MH:
multi-hop, CM: cross-modal, PU: person understanding, GK: general knowledge. Overall
scores are reported in Table~\ref{tab:main1}.}
\label{tab:m3breakdown}
\begin{tabular}{l|ccccc|ccccc}
\toprule
& \multicolumn{5}{c|}{M3-bench-robot} & \multicolumn{5}{c}{M3-bench-web} \\
Method & ME & MH & CM & PU & GK & ME & MH & CM & PU & GK \\
\midrule
Qwen2.5-VL-7B       & 2.9 & 3.8 & 3.6 & 4.6 & 3.4 & 11.9 & 10.5 & 13.4 & 14.0 & 20.9 \\
Gemini-1.5-Pro      & 6.5 & 7.5 & 8.0 & 9.7 & 7.6 & 18.0 & 17.9 & 23.8 & 23.1 & 28.7 \\
GPT-4o              & 9.3 & 9.0 & 8.4 & 10.2 & 7.3 & 21.3 & 21.9 & 30.9 & 27.1 & 39.6 \\
MovieChat           & 13.3 & 9.8 & 12.2 & 15.7 & 7.0 & 12.2 & 6.6 & 12.5 & 17.4 & 11.1 \\
Flash-VStream       & 21.6 & 19.4 & 19.3 & 24.3 & 14.1 & 24.5 & 10.3 & 24.6 & 32.5 & 20.2 \\
Gemini-Agent        & 15.8 & 17.1 & 15.3 & 20.0 & 15.5 & 29.3 & 20.9 & 33.8 & 34.6 & 45.0 \\
M3-Agent            & 32.8 & 29.4 & 31.2 & 43.3 & 19.1 & 45.9 & 28.4 & 44.3 & 59.3 & 53.9 \\
GPT-5               & 34.7 & 35.3 & 32.1 & 43.3 & 32.1 & 55.5 & 50.1 & 55.8 & 61.8 & 46.9 \\
\midrule
ReflectWorld-MM     & 38.0 & 35.3 & 37.4 & 52.4 & 25.1 & 53.2 & 48.3 & 53.8 & 59.8 & 59.3 \\
\bottomrule
\end{tabular}
\end{table*}

\begin{table}[t]
\centering
\footnotesize
\setlength{\tabcolsep}{5pt}
\caption{EgoLife-QA per-sub-task breakdown (accuracy, \%). Gemini-1.5-Pro is quoted from
the EgoLife paper~\citep{egolife}. GPT-5 and ReflectWorld-MM are our current runs. EL:
EntityLog, ER: EventRecall, HI: HabitInsight, RM: RelationMap, TM: TaskMaster.}
\label{tab:egobreakdown}
\begin{tabular}{lcccccc}
\toprule
Model & EL & ER & HI & RM & TM & Avg \\
\midrule
Gemini-1.5-Pro      & 36.0 & 37.3 & 45.9 & 30.4 & 34.9 & 36.9 \\
GPT-5               & 49.6 & 45.2 & 47.5 & 33.6 & 36.5 & 42.6 \\
\midrule
ReflectWorld-MM     & 52.9 & 51.6 & 38.3 & 40.8 & 51.9 & 46.8 \\
\bottomrule
\end{tabular}
\end{table}

\section{Additional Implementation Details}
\label{app:implementation}
\noindent\textbf{Models and store.} Perception uses a configurable vision--language model
(GPT-5-mini in our experiments), and the memory service uses GPT-5-mini for consolidation;
memory items are embedded with OpenAI's \texttt{text-embedding-3-small}
($1{,}536$ dimensions). The store is a vector
database with separate collections for the three episodic levels, the entity-centric and
per-camera semantic memories, the procedural rules, and the face and body
re-identification galleries. A reserved visual-embedding collection also exists but is not
on the text-search path, so the three episodic \emph{levels} in the main text refer to the
abstraction hierarchy rather than to the full set of collections.

\noindent\textbf{Local perception utilities.} The deployment stack also includes a small
set of open-source perception models converted to ONNX. These models are used as local
utilities that provide bounded visual or audio signals to the memory pipeline; they are not
the main reasoning model. The released code documents the expected ONNX artifacts; where
upstream licenses permit, converted artifacts can be distributed, and otherwise users can
download the corresponding open-source weights and convert them, or replace any module with a
stronger compatible model.
Table~\ref{tab:onnxmodels} summarizes the
capabilities used by our default local configuration.

\begin{table*}[t]
\centering
\footnotesize
\setlength{\tabcolsep}{5pt}
\caption{Local perception utilities used by the deployment stack. Required modules are
used by the default local perception profile; optional modules can be enabled or replaced.}
\label{tab:onnxmodels}
\begin{tabular}{>{\raggedright\arraybackslash}p{0.18\textwidth}
                >{\raggedright\arraybackslash}p{0.30\textwidth}
                >{\raggedright\arraybackslash}p{0.40\textwidth}}
\toprule
Capability & Default model & Role \\
\midrule
Object detection & YOLO26m & General object detection; required. \\
Face detection & RetinaFace-ResNet50 & Face boxes and landmarks; required. \\
Face ReID & ArcFace-MobileFaceNet & Face embedding; required. \\
Body ReID & CLIP-ReID ViT-B & Body embedding for optional experiments. \\
Body pose & RTMPose-S & 17-point pose for optional experiments. \\
Speech recognition & Moonshine Tiny int8 via sherpa-onnx & Optional local STT; cloud STT can be used instead. \\
\bottomrule
\end{tabular}
\end{table*}

\noindent\textbf{Episodic writes.} The three episodic levels follow the autobiographical
hierarchy of event-specific knowledge, general events, and lifetime periods. Each segment
writes one entity-level observation per resolved entity, keyed by the persistent entity
identifier and storing appearance, behavior, and interactions at that moment. The trace
level writes one segment summary and links back to the participating entity observations.
The schema level accumulates closed events and, once a budget of events or a time interval
is reached, consolidates them into a chapter-level summary that indexes the underlying
traces.

\noindent\textbf{Perception scheduling and anchoring.} The adaptive controller couples a
deterministic frame-level policy with a semantic segment-level policy. A strategy selector
chooses the analysis level---a lightweight vision-only level, or a fuller level that adds
local detection and re-identification---from the capture mode and the locally installed
models. After gateway normalization, the stream is split at natural boundaries using
visual motion and voice activity; each segment carries its frames and a transcript, with
speaker diarization when available. A gate skips analysis on static or silent segments, but
hard guards always analyze the first segment of an event, any segment containing speech, and
any segment whose audio could not be transcribed. Vision--language calls carry per-stage
budgets such as output length and reasoning effort. For anchoring, every detection in a segment receives a
local anchor, and the model may attach an identity target only to an existing anchor; a pure
parser enforces this together with a small set of consistency rules before any target updates
the memory.

\noindent\textbf{Context layers.} The three context layers that enhance perception are
realized as follows. The working memory is kept per active event and holds a rolling event
summary (capped at about one hundred words), the three most recent segment summaries, the
active entities, and up to eight tracked targets; a target is dropped after a few consecutive
missed segments, an entity after an inactivity timeout of a few minutes, and the whole event
is rotated once it exceeds a maximum number of segments. The per-camera semantic context
(scene type, typical routines, and an anomaly baseline) is maintained by an LLM that revises
it periodically once enough observations have accumulated, and it is rendered into the
perception prompt as scene and routine directives. The entity-history layer is populated on a
re-identification hit: the matched entity's recent episodic and semantic records are retrieved
and added to the prompt as continuity hints.

\noindent\textbf{Agent steering.} The high-level agent acts through a vision controller with a
deterministic frame-level policy and a semantic segment-level policy. When a segment looks
risk-relevant the controller may consult the host agent, subject to a cooldown of about one
minute of media time and a small per-event budget; the reply is restricted to an allowlist---a
scene description, focus targets, and security rules---that is injected into subsequent
perception prompts, and it must carry an explicit exit condition. In deployment this
consultation is served by the host assistant; the benchmark path uses a fixed policy.

\noindent\textbf{Identity.} Re-identification matches face and body embeddings against
multi-slot galleries and returns scored candidates; the resolver alone commits an entity
identifier, and evidence flagged unsafe may be displayed but never writes a gallery, name,
or identity. Cross-segment continuity is carried by the short-term memory and weak pending
evidence rather than by a tracker.

\noindent\textbf{Consolidation.} Entity-centric semantic memory is consolidated every
$N{=}5$ observations of an entity. The per-entity counter is persistent and scoped per
user, the current segment is excluded as self-evidence, and the merge is skipped when an
entity has no prior history. Importance is reinforced with growth rate $\gamma{=}0.2$ and
capped at one. For \textsc{Update} and \textsc{Delete}, the consolidator must point to an
existing semantic fact identifier; edits without a valid target are dropped. Identity
facts are written at maximum importance and are protected from automatic update or
deletion.

\noindent\textbf{Streaming commit.} The implementation exposes two execution modes:
\texttt{offline\_quality}, used for deterministic benchmarks, and \texttt{live\_latency},
used for live operation. In \texttt{live\_latency}, the identity-critical fields of a
segment (event boundary, identity, matched rules, summary) are committed under a per-user
lock as soon as they are validated, releasing the next segment, after which the enrichment
fields reconcile against the frozen critical state. Full prompts and exact thresholds are
provided with the released code.

\section{Agent and Memory Access}
\label{app:agent}
The read side of ReflectWorld-MM is an agent that answers questions by calling memory
tools. The perception and memory subsystem is independent of which agent is used.

\noindent\textbf{Benchmark agent.} For the benchmarks, the agent is a tool-calling
question-answering agent driven by GPT-5. It is given three tools over the memory store: a
similarity \emph{search} over a chosen memory level; a \emph{get} that returns an item
together with its time-ordered neighbors; and a \emph{visual-evidence} tool that fetches the
key frames behind a memory item, so the agent can confirm a detail visually rather than rely
on the stored text alone. The agent issues these calls over several turns, accumulating
evidence until it can answer, and then returns its answer. Multiple-choice answers are scored
by option matching, and open-ended M3-bench answers are scored by a GPT-5-mini judge.

\noindent\textbf{Deployment via OpenClaw.} In deployment, ReflectWorld-MM is packaged as a
plugin for an off-the-shelf assistant runtime (OpenClaw) through a tool-based contract. It
registers a small, uniform set of tools that expose capture control, perception, memory
query, notification policy, and per-camera context to the host assistant. The assistant can
therefore start watching a new source, ask what is happening, query the long-term memory, or
set a rule, all through ordinary tool calls and without any change to the underlying system.
The benchmark agent and the deployed assistant thus use the same memory interface; only the
surrounding agent differs.

\section{Connecting Arbitrary Video Streams}
\label{app:streams}
ReflectWorld-MM ingests an arbitrary live source through a common gateway. Supported sources
include network cameras (RTSP/RTSPS), local video files, USB webcams, HTTP streams, and
smartphone cameras. A live media gateway normalizes each source into a uniform stream that
the capture stage reads, so the rest of the pipeline is independent of where the video comes
from. For a smartphone, a lightweight bridge turns the phone's browser camera into a local
stream over WebRTC, which the gateway then serves like any other source. Because ingestion is
decoupled from perception and memory, supporting a new source type requires only a gateway
adapter, and nothing downstream changes.

\section{The ReflectWorld-MM Dashboard}
\label{app:dashboard}
\begin{figure*}[t]
\centering
\includegraphics[width=0.92\textwidth]{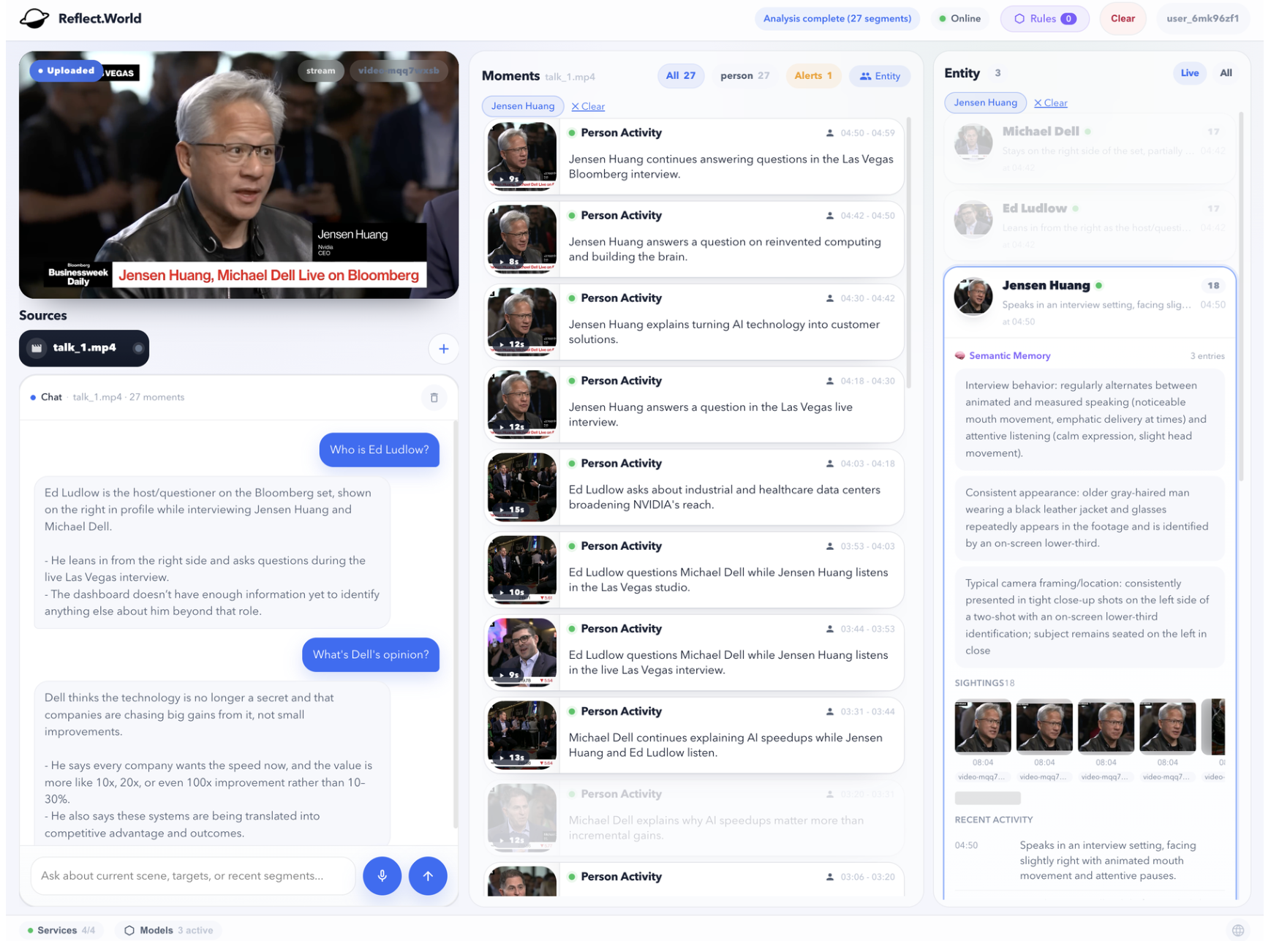}
\caption{The ReflectWorld-MM dashboard on a live interview. Left: the video source and a chat
panel where an agent answers from memory (e.g., ``Who is Ed Ludlow?''). Middle: per-segment
moments with timestamps. Right: the entity panel for one person, with the evolving semantic
memory, sightings, and recent activity written during perception.}
\label{fig:dashboard}
\end{figure*}

We built a dashboard for the system, shown in Figure~\ref{fig:dashboard} on a live
interview. On the left are the video source and a chat
panel where the agent answers questions from memory. In the middle are the per-segment
moments, that is, entity-resolved activities with timestamps. On the right is the entity
panel, with each entity's evolving semantic memory, its sightings, and its recent activity.
The dashboard is a window onto the same memory the benchmark agent queries; it is not a
separate system.

\section{Qualitative Benchmark Traces}
To show that the agent answers from memory written \emph{during} perception rather than by
re-watching the video, we walk through four benchmark questions. Each figure lays out the
question, the memory that was already stored when the question arrived, how the agent used that
memory, and the answer.

\begin{figure*}[t]
\centering
\includegraphics[width=\textwidth]{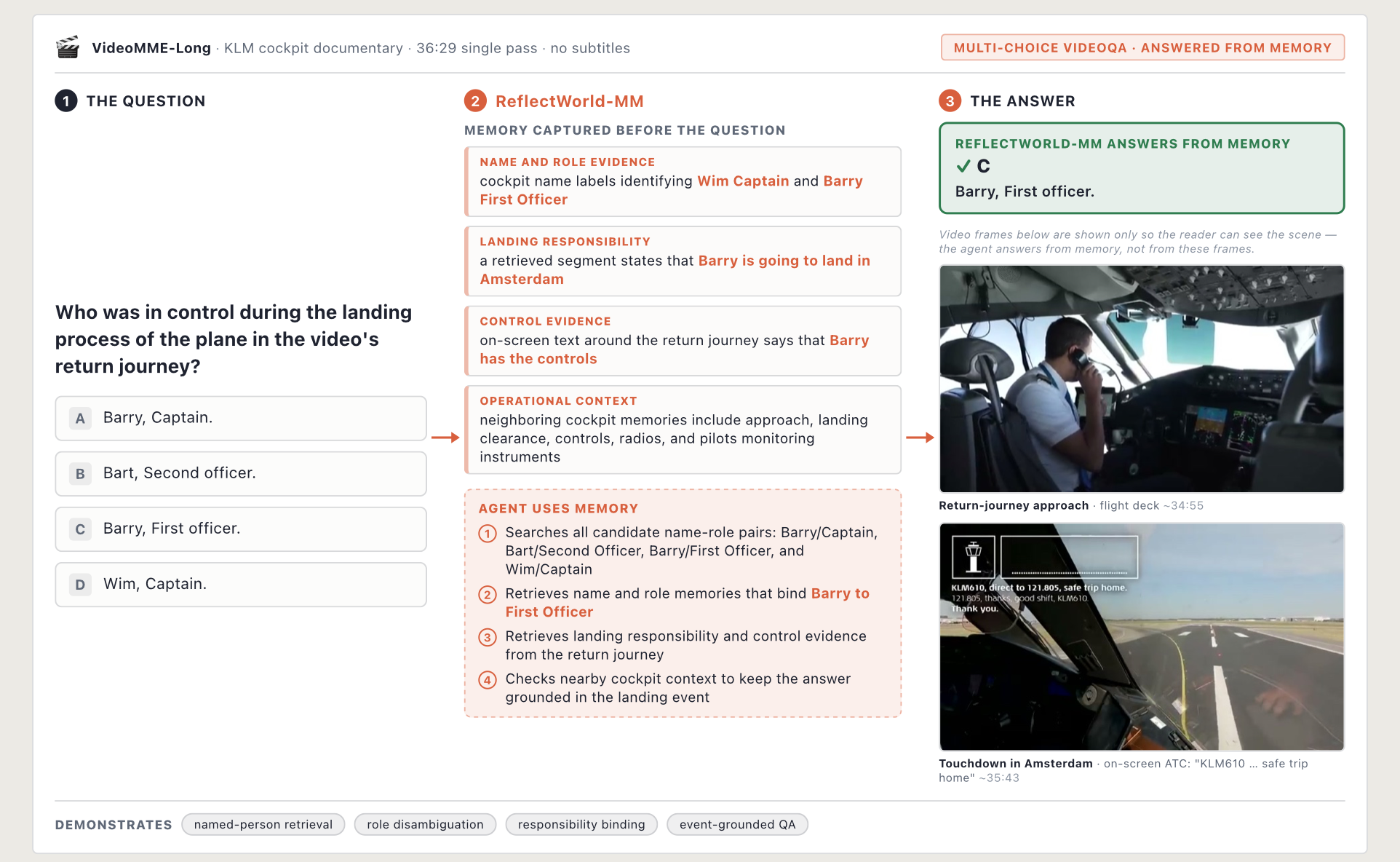}
\caption{Qualitative trace on VideoMME (long). The role question is answered entirely from
memory written during perception---cockpit name--role bindings, the landing-responsibility
evidence, and the operational context---which the agent retrieves and grounds in the landing
event to answer ``Barry, First Officer.''}
\label{fig:case2}
\end{figure*}

\noindent\textbf{VideoMME (long): named person, role, and responsibility
(Figure~\ref{fig:case2}).} On a long cockpit documentary, the question asks who controlled the
landing on the return journey. By then, perception has bound the cockpit name labels to roles
(Wim as captain, Barry as first officer) and recorded who held the controls during the approach.
The agent searches the candidate name--role pairs, retrieves the binding of Barry to first
officer together with the landing-responsibility evidence, and grounds them in the landing event
to answer ``Barry, First Officer.'' This trace exercises named-person retrieval, role
disambiguation, responsibility binding, and event-grounded question answering.

\begin{figure*}[t]
\centering
\includegraphics[width=\textwidth]{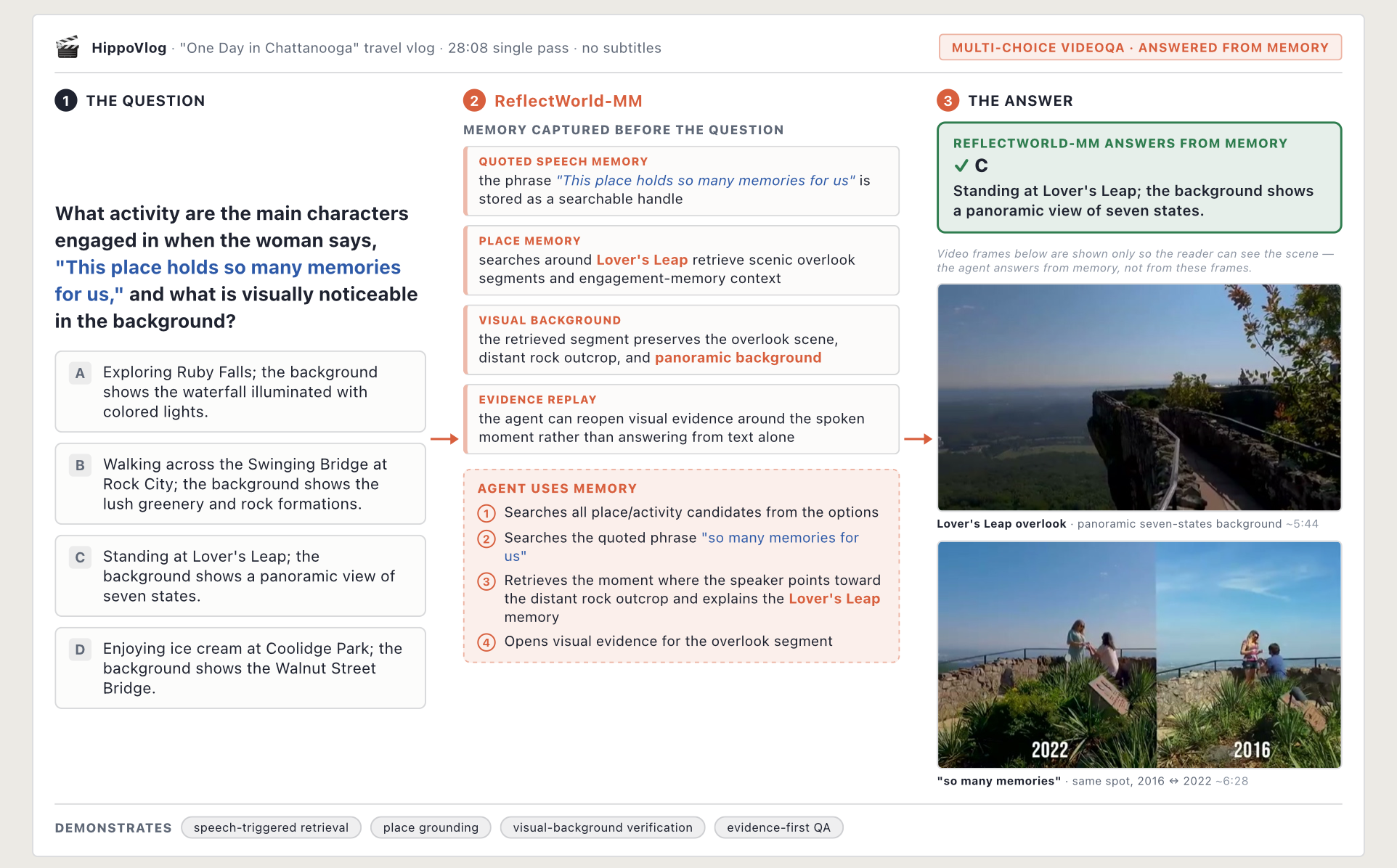}
\caption{Qualitative trace on HippoVlog. The speech-grounded place question is answered from
memory: the quoted phrase, the place (Lover's Leap), and its panoramic background were all
stored during perception, and the agent retrieves the spoken moment and verifies the background.}
\label{fig:case3}
\end{figure*}

\noindent\textbf{HippoVlog: speech-triggered place grounding (Figure~\ref{fig:case3}).} On a
travel vlog, the question pairs a spoken phrase (``this place holds so many memories for us'')
with the activity and the visible background at that moment. Perception has stored the quoted
phrase as a searchable handle and preserved the place (Lover's Leap) together with its panoramic
background. The agent retrieves the quoted phrase, locates the overlook moment, and verifies the
visual background, answering that the characters are standing at Lover's Leap with a panoramic
view of seven states. This trace exercises speech-triggered retrieval, place grounding,
visual-background verification, and evidence-first question answering.

\begin{figure*}[t]
\centering
\includegraphics[width=\textwidth]{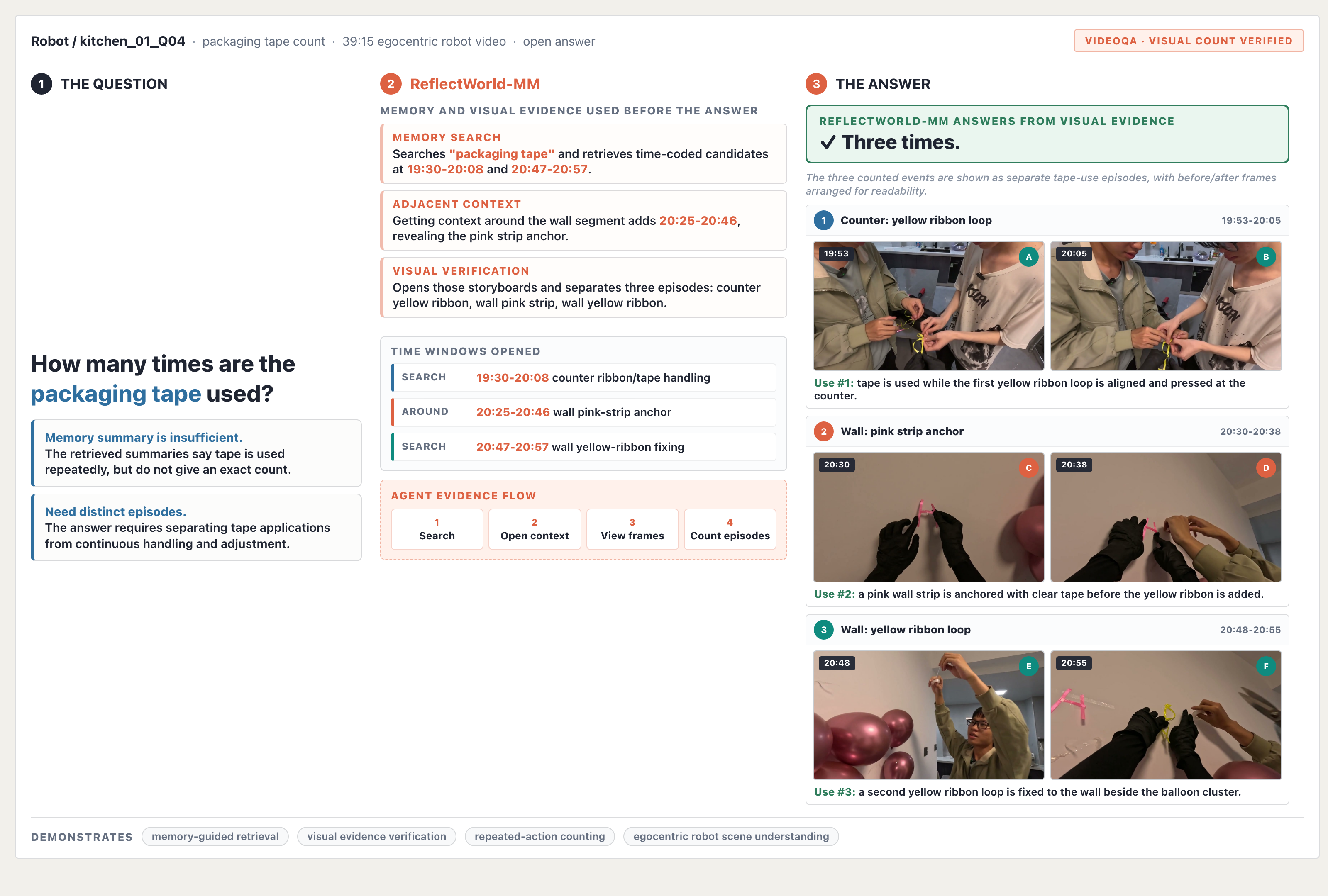}
\caption{Qualitative trace on M3-bench-robot. The tape-counting question cannot be answered
from a coarse summary alone; the agent retrieves candidate tape-use windows, opens adjacent
context, verifies the frames, and counts three distinct tape-use episodes.}
\label{fig:case_m3_tape}
\end{figure*}

\noindent\textbf{M3-bench-robot: repeated-action counting with visual verification
(Figure~\ref{fig:case_m3_tape}).} On an egocentric robot video, the question asks how many
times packaging tape is used. The retrieved memory summary alone is insufficient, because it
indicates repeated tape handling but does not give an exact count. The agent therefore uses
memory search to find candidate tape-use windows, opens adjacent context around the wall
segment, and verifies the relevant frames. It separates three episodes: a yellow ribbon loop
at the counter, a pink strip anchor on the wall, and a second yellow ribbon loop fixed to the
wall. This trace exercises memory-guided retrieval, adjacent-context expansion, visual
evidence verification, and repeated-action counting.

\begin{figure*}[t]
\centering
\includegraphics[width=\textwidth]{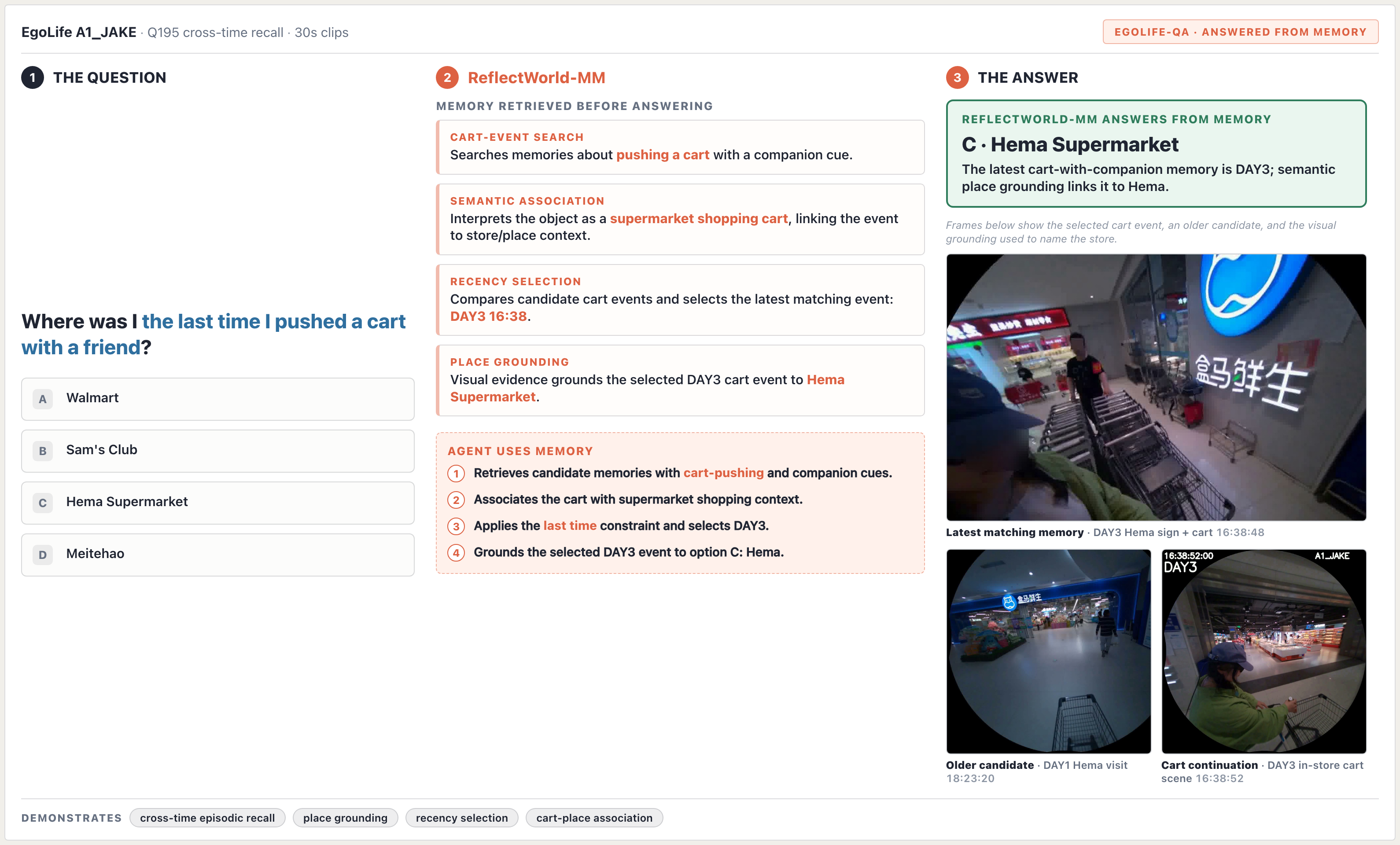}
\caption{Qualitative trace on EgoLife-QA. The agent retrieves candidate cart-pushing memories,
selects the latest matching event, and grounds the selected event to Hema Supermarket using
visual place evidence.}
\label{fig:case_egolife_cart}
\end{figure*}

\noindent\textbf{EgoLife-QA: cross-time location recall
(Figure~\ref{fig:case_egolife_cart}).} On an egocentric life log, the question asks where the
person was the last time they pushed a cart with a friend. The agent searches memories about
cart pushing with a companion cue, interprets the object as a supermarket shopping cart, and
compares candidate cart events across time. It selects the latest matching event on DAY3 and
grounds the place with visual evidence of the Hema sign, answering ``Hema Supermarket.'' This
trace exercises cross-time episodic recall, recency selection, semantic place association, and
visual place grounding.

\end{document}